\documentclass[pdflatex,sn-mathphys-num]{sn-jnl}

\usepackage{graphicx}
\usepackage{amsmath,amssymb,amsfonts}
\usepackage{booktabs}
\usepackage{array}
\usepackage[table]{xcolor}
\usepackage{textcomp}
\usepackage{placeins}

\newcolumntype{L}[1]{>{\raggedright\arraybackslash}p{#1}}
\definecolor{reviewboxbg}{HTML}{F1F0E6}
\definecolor{tableheadbg}{HTML}{E7E4D8}
\definecolor{methodrowone}{HTML}{F8F6ED}
\definecolor{methodrowtwo}{HTML}{EEF3F4}
\definecolor{roadmapblue}{HTML}{214EAF}
\definecolor{roadmapgreen}{HTML}{2D7A36}
\definecolor{roadmaporange}{HTML}{D46A1F}
\definecolor{roadmappurple}{HTML}{7A3FA2}

\begin{document}

\title[Medical world models]{Medical world models: representing medical states, modelling clinical dynamics and guiding intervention policies}

\author[1]{\fnm{Ke} \sur{Liu}}\email{ke.liu@zju.edu.cn}
\equalcont{These authors contributed equally to this work.}
\author[1]{\fnm{Mengxuan} \sur{Li}}\email{limx97@zju.edu.cn}
\equalcont{These authors contributed equally to this work.}
\author[1]{\fnm{Yanyi} \sur{Bao}}\email{baoyanyi@tju.edu.cn}
\author[2]{\fnm{Tianyun} \sur{Zhang}}\email{12218151@zju.edu.cn}
\author[3]{\fnm{Chong} \sur{Chu}}\email{chong\_chu@harvard.edu}
\author[1]{\fnm{Jiajun} \sur{Bu}}\email{bjj@zju.edu.cn}
\author*[1]{\fnm{Haishuai} \sur{Wang}}\email{haishuai.wang@zju.edu.cn}

\affil*[1]{\orgdiv{College of Computer Science}, \orgname{Zhejiang University}, \orgaddress{\city{Hangzhou}, \country{China}}}
\affil[2]{\orgdiv{School of Medicine}, \orgname{Zhejiang University}, \orgaddress{\country{China}}}
\affil[3]{\orgdiv{Department of Biomedical Informatics}, \orgname{Harvard University}, \orgaddress{\country{USA}}}

\abstract{Medical diagnosis and treatment are dynamic processes in which patient states evolve over time and clinical interventions alter future outcomes. Although current medical AI can detect disease, estimate risk and generate reports, many systems still return static labels or scores, offering limited insight into how illness may progress or how alternative interventions may reshape its trajectory. Medical world models adapt the world-model idea from artificial intelligence to healthcare by learning internal simulators of patient-state dynamics. Their long-term goal is to help clinicians anticipate deterioration, compare treatment-conditioned futures and tailor care to individual patients. Yet relevant work remains scattered across foundation models, longitudinal modelling, disease simulation, treatment-effect estimation, reinforcement learning and digital twins. To bridge this gap, this review outlines a roadmap for advancing medical AI from isolated diagnosis and prediction toward medical world models that simulate disease evolution and support intervention decisions. This roadmap is organized around three coupled capabilities: patient-state construction, clinical dynamics modelling and intervention decision support. Across representative systems, the comparison highlights what each capability contributes and how partial components can be integrated into more mature perception--dynamics--planning systems. Finally, we identify the challenges involved in turning plausible rollouts into clinically useful simulators. Related literature is available at \href{https://github.com/1999kevin/awesome_medical_world_models}{1999kevin/awesome\_medical\_world\_models}.}


\maketitle

\section{Introduction}
\label{sec:introduction}

Medical diagnosis and treatment are fundamentally processes of reasoning about and influencing evolving patient states. Rather than making decisions from isolated observations, clinicians continuously integrate symptoms, medical history, physiological measurements, laboratory results and medical images to infer a patient's current condition. This inferred state changes over time as disease progresses, interventions take effect and new evidence becomes available. Clinical decision-making therefore extends beyond recognizing the present condition to anticipating future trajectories and selecting interventions that may alter them. Diagnosis, prognosis and intervention planning are thus tightly coupled in an iterative process in which patient states evolve over time and clinical interventions shape future outcomes \cite{moor_foundation_2023,shmatko_natural_2025,komorowski_artificial_2018}.

Medical AI has made important parts of this workflow computable. Deep learning and foundation models can detect disease, learn reusable patient representations and support tasks such as report generation and medical question answering \cite{lecun2015deep,miotto_deep_2016,bommasani2021opportunities,moor_foundation_2023}. Yet many systems still return static labels, risk scores or generated samples. They often do not specify how illness may progress over time, how uncertainty accumulates during rollout or how alternative interventions would reshape the same patient's trajectory.

World models offer a way to make the missing temporal structure explicit. In artificial intelligence, a world model learns an internal state and dynamics model that can be used to imagine future states before acting \cite{sutton1998reinforcement,ha_world_2018,hafner2019learning,hafnerdream}. Modern neural implementations make this idea practical by learning compact latent states rather than modelling every detail of the raw observation. \hyperref[box:world-model-definition]{Box~1} summarizes the general concept and its relationship to adjacent paradigms. This view is especially relevant to medicine, where the useful state is often not a raw image, waveform or record, but a representation of anatomy, physiology, disease burden, history, uncertainty and treatment context.

\begin{table}[t]
\phantomsection\label{box:world-model-definition}
\centering
\begingroup
\setlength{\fboxsep}{0pt}
\colorbox{reviewboxbg}{%
\begin{minipage}{0.98\columnwidth}
\vspace{0.6em}
\hspace{0.05\linewidth}
\begin{minipage}{0.88\linewidth}
{\small\bfseries BOX 1}
\vspace{0.25em}

\hrule
\vspace{0.4em}

{\large\bfseries What is a world model?}
\vspace{0.3em}

\hrule
\vspace{0.65em}

\small
World models are internal models that learn how an environment evolves and how actions change future states \cite{sutton1998reinforcement,ha_world_2018,hafner2019learning,hafnerdream}. At their core, they can be defined as action-conditioned transition models over states, $p(s_{t+1}\mid s_t,a_t)$, where $s_t$ denotes the current state and $a_t$ denotes an action. The state need not be a raw observation. It can be a compact latent representation that preserves information needed for future prediction and control. Modern neural world models often organize this core around observations $o_t$, latent states $s_t$, actions $a_t$, a transition or dynamics model $p(s_{t+1}\mid s_t,a_t)$ and, when the model is used for planning or training, an objective or cost function that scores future states or prediction errors \cite{sutton1998reinforcement,ha_world_2018,hafner2019learning,hafnerdream}.

World models are useful when an agent must imagine possible futures before acting. In autonomous driving, a model can roll out how a traffic scene may evolve under candidate manoeuvres. In robotics, it can predict how pushing, grasping or navigation actions may change object configurations, contacts and goals \cite{finn_deep_2017,zhao_vlmpc_2024,sanchez-gonzalez_graph_2018}. In spatial intelligence, video generation, games and interactive environments, related models infer 3D structure, learn action-controllable worlds or support planning inside learned latent environments \cite{ha_world_2018,hafner2019learning,hafnerdream,bruce_genie_nodate}.

World models differ from adjacent technologies by their functional role. Foundation models provide reusable representations, but may not support rollouts. Generative models synthesize plausible samples, but realism alone does not imply action-conditioned dynamics. Digital twins emphasize high-fidelity replicas of specific systems, whereas world models can be learned, latent and task-oriented. Causal models estimate interventions and counterfactuals, but may not provide long-horizon simulated trajectories. Model-free policies choose actions without necessarily exposing an inspectable model of future states.
\end{minipage}
\vspace{0.7em}
\end{minipage}}
\endgroup
\end{table}

For a medical world model, the corresponding environment is the evolving patient and the surrounding care process. The central modelling problem can be written abstractly as
\begin{equation}
    p(s_{t+1}\mid s_t,a_t),
    \label{eq:transition}
\end{equation}
where $s_t$ denotes a latent patient state and $a_t$ denotes an action, intervention, procedure, monitoring choice, exposure or control signal. The state may be inferred from images, EHRs, physiological waveforms, reports, molecular measurements or multimodal observations. The next state may appear as future clinical events, imaging changes, physiological trajectories, tumour evolution, procedure states or counterfactual outcomes. This abstraction connects clinical prediction, treatment-effect reasoning, reinforcement learning and digital twins through a shared state-transition view \cite{rubin2005causal,komorowski_artificial_2018,laubenbacher_digital_twins_2024}. Yet relevant work remains scattered across patient representations, longitudinal EHR modelling, generative imaging, disease simulation, treatment-effect estimation, digital twins, offline reinforcement learning and embodied medical AI \cite{miotto_deep_2016,kraljevic_foresightgenerative_2024,yang2025medical,ding_clarity_2026,xu2026meddreamer}. The field therefore lacks a unified roadmap for judging which methods construct clinically useful states, which simulate plausible futures and which can support intervention decisions.

\begin{figure*}[t]
\centering
\includegraphics[width=\textwidth]{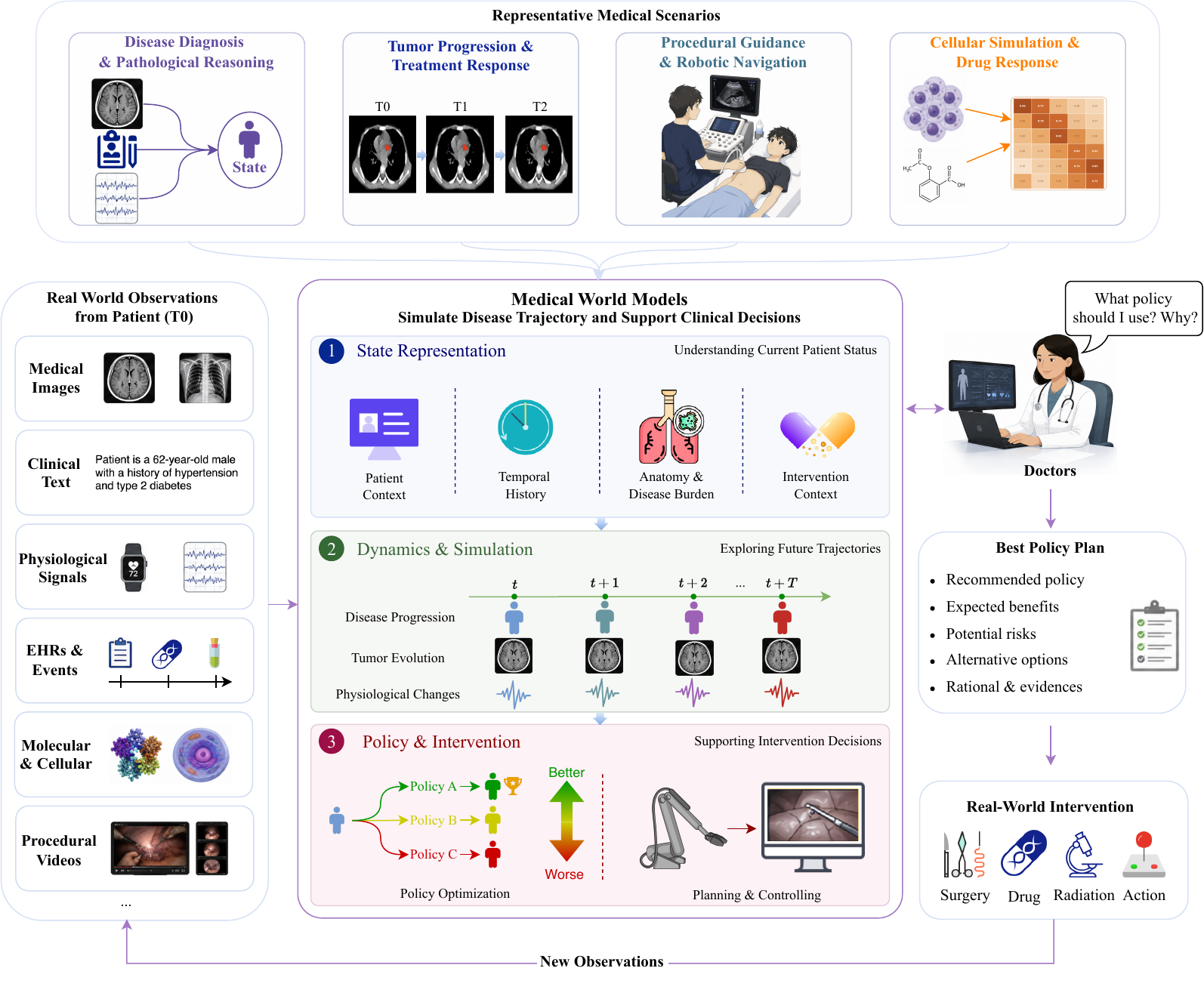}
\caption{Medical world models as a unifying computational abstraction. Diverse medical settings, from patient-state reasoning and tumour-response simulation to procedural guidance and cellular perturbation modelling, can be organized around the same computational problem: constructing a clinically useful state, modelling clinical dynamics and using simulated futures to guide intervention.}
\label{fig:loop}
\end{figure*}

This review addresses this gap by developing a roadmap for advancing medical AI from isolated diagnosis and prediction toward medical world models that simulate disease evolution and support intervention decisions (Fig.~\ref{fig:loop}). It makes three contributions. First, it translates the general world-model formulation into a clinical simulator problem centred on patient-state construction, clinical dynamics modelling and intervention decision support. Second, it reorganizes fragmented work across foundation models, longitudinal modelling, generative imaging, biological simulation, digital twins, causal inference, reinforcement learning and embodied medical AI according to the role each method plays in this simulator loop. Third, it distinguishes partial components from more mature perception--dynamics--planning systems and identifies the evidence needed to turn plausible rollouts into clinically useful simulators.

\section{Medical world models as clinical simulators}
\label{sec:clinical-simulators}

A medical world model is best understood as an internal simulator of patient states and their possible futures. This definition is functional rather than architectural. It does not require one specific model family, data modality or clinical task. It follows the broader world-model view in which agents learn state transitions and use imagined futures to support action selection \cite{sutton1998reinforcement,ha_world_2018,hafner2019learning,hafnerdream}. In clinical settings, the generic terms of a world model map onto three coupled capabilities: patient-state construction, clinical dynamics modelling and intervention decision support (Fig.~\ref{fig:roadmap-taxonomy}). The terminology used for these components is collected in Appendix~\ref{app:key-concepts}.

\begin{figure}[!t]
\centering
\includegraphics[width=\textwidth]{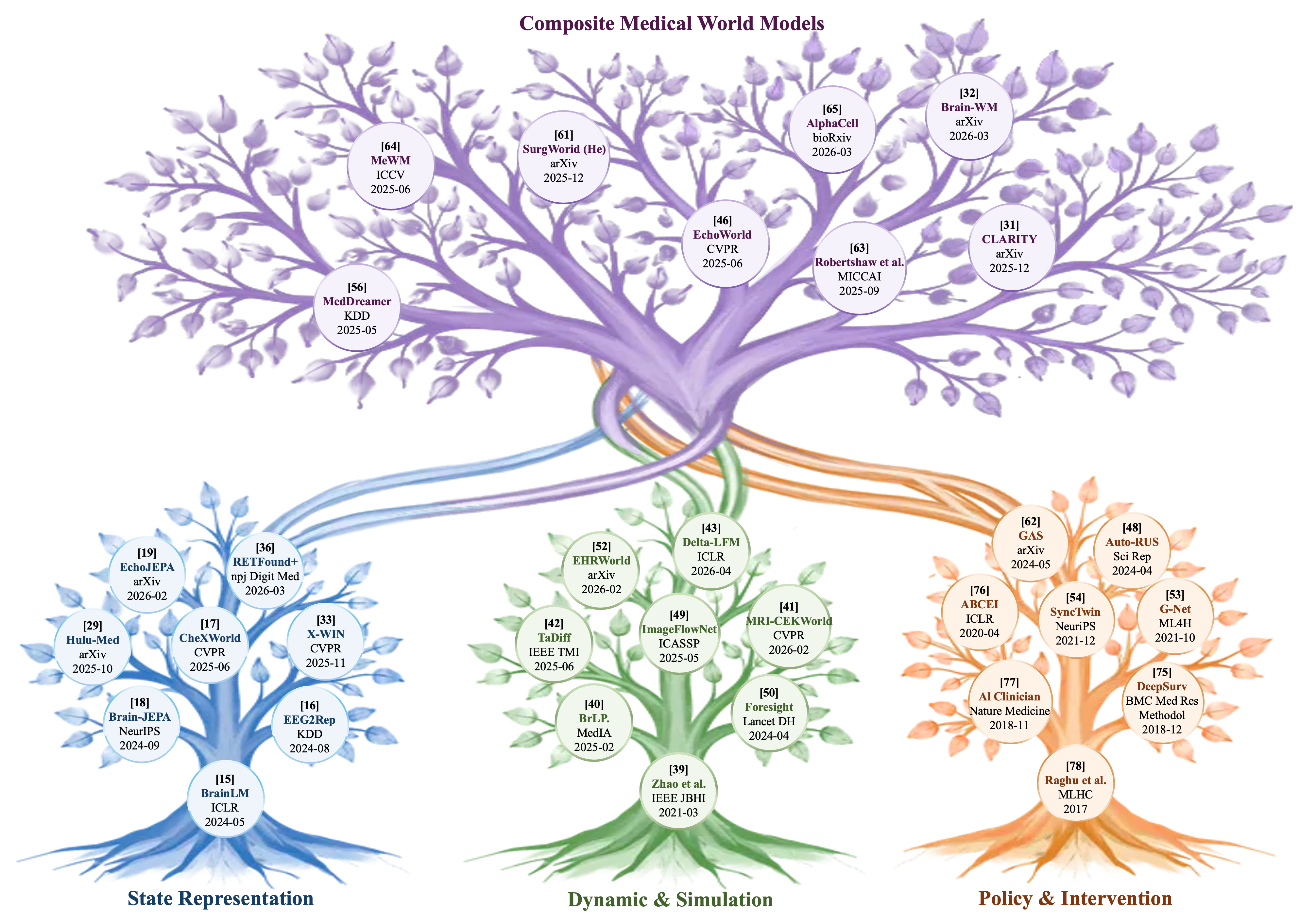}
\caption{Roadmap to medical world models. The field can be organized into three continuing technical streams: \textcolor{roadmapblue}{transition-ready state representation}, \textcolor{roadmapgreen}{disease dynamics simulation} and \textcolor{roadmaporange}{intervention decision support}. These streams continue to develop independently, while \textcolor{roadmappurple}{composite Medical World Models} begin to integrate selected capabilities into patient-state simulators that support prediction, rollout and intervention comparison.}
\label{fig:roadmap-taxonomy}
\end{figure}

\begin{table*}[!t]
\caption{Representative medical world-model methods and adjacent predictive simulators by clinical modality.}
\label{tab:modality-architecture-applications}
\fontsize{6.3pt}{6.6pt}\selectfont
\setlength{\tabcolsep}{2.5pt}
\renewcommand{\arraystretch}{0.82}
\begin{tabular}{@{}L{0.14\textwidth}L{0.30\textwidth}L{0.50\textwidth}@{}}
\toprule
\rowcolor{tableheadbg}
Modality & Method family in the medical world-model loop & Applications and representative studies \\
\midrule
\rowcolor{methodrowone}
X-ray & Self-supervised radiograph state representation & Chest radiograph representation learning and transition-ready visual features \cite{yue2025chexworld}. \\
\rowcolor{methodrowtwo}
 & Cross-view and 3D-aware X-ray modelling & Cross-view prediction, depth-aware sensing and 3D-aware radiograph modelling \cite{yang2025x,yang_xray2xray_2025}. \\
\midrule
\rowcolor{methodrowone}
CT & Cross-modal semantic alignment & Automated 3D CT report generation and zero-shot CT interpretation \cite{hamamci_ct2rep_2024,park_radzero3d_nodate}. \\
\rowcolor{methodrowtwo}
 & Longitudinal image registration and outcome modelling & Deformable image trajectories for adaptive radiotherapy and dynamic contrast-enhanced CT analysis for survival and surgical-margin prediction \cite{lee_seq2morph_2023,yao_deepprognosis_2021}. \\
\midrule
\rowcolor{methodrowone}
MRI & Brain-dynamics foundation modelling & fMRI state representation and JEPA-style brain dynamics foundation models \cite{caro_brainlm_2023,dong_brain-jepa_2024}. \\
\rowcolor{methodrowtwo}
 & Image-trajectory and disease-progression modelling & Brain disease progression, glioma growth and patient-specific longitudinal image generation \cite{zhao_prediction_2021,puglisi_brain_2025,kong2026mri}. \\
\rowcolor{methodrowone}
 & Treatment-aware generative dynamics & Post-treatment prediction, contrast-enhancement kinetics and treatment-response simulation \cite{liu_treatment-aware_2025,chen_learning_2026,leclercq_pre_2026}. \\
\midrule
\rowcolor{methodrowone}
Ultrasound & Structure-constrained and JEPA-style latent state learning & Echocardiography representation learning and ultrasound-video segmentation \cite{jiang2024structure,munim2026echojepa,ellis_self-supervised_2025}. \\
\rowcolor{methodrowtwo}
 & Procedural dynamics and embodied planning & Automatic probe guidance, probe-motion planning and robotic ultrasound path planning \cite{yue2025echoworld,jiang_cardiac_2024,bi_autonomous_2024}. \\
\midrule
\rowcolor{methodrowone}
Retina & Long-context retinal state modelling & Disease incidence prediction, risk stratification and progression-aware retinal representations \cite{wang_time_2026}. \\
\rowcolor{methodrowtwo}
 & Irregular image-trajectory forecasting & Multiscale retinal disease progression forecasting from sparse longitudinal images \cite{liu_imageflownet_2025}. \\
\midrule
\rowcolor{methodrowone}
Electronic health records & Self-supervised patient-state representation & Latent patient phenotyping and reusable EHR state representations \cite{miotto_deep_2016}. \\
\rowcolor{methodrowtwo}
 & Generative event-trajectory modelling & Patient timeline generation and long-horizon clinical event forecasting \cite{kraljevic_foresightgenerative_2024,waxler_generative_2025,mu_ehrworld_2026}. \\
\rowcolor{methodrowone}
 & Policy learning with latent imagination & Counterfactual trajectory estimation, treatment-effect modelling, EHR-based latent rollouts, AKI intervention optimization and decision-support simulations \cite{li_g-net_nodate,qian_synctwin_nodate,adam2026patient,xu2026meddreamer,zhang2026world}. \\
\midrule
\rowcolor{methodrowone}
Physiological signals and wearables & JEPA-style latent state learning & EEG representation learning with video-style predictive embedding objectives \cite{hojjati_video_2026}. \\
\rowcolor{methodrowtwo}
 & Masked and time-series self-supervised modelling & ECG representation learning, behavioural time-series modelling and downstream diagnostic prediction \cite{weimann_self-supervised_2025,mohammadi_foumani_eeg2rep_2024,xie_jets_nodate}. \\
\midrule
\rowcolor{methodrowone}
Surgical video and robotics & JEPA-style surgical state representation & Cataract surgery representation learning and action-conditioned procedural prediction \cite{shah_vision_nodate,koju_surgical_2026}. \\
\rowcolor{methodrowtwo}
 & Procedural trajectory and scene dynamics & Surgical scene forecasting, suturing dynamics and procedure-state prediction \cite{turkcan2025towards,shah2026learning,he2025surgworld}. \\
\rowcolor{methodrowone}
 & Embodied planning and control & Surgical grasping, robot policy learning and world-model-based control \cite{lin_world_2024}. \\
\midrule
\rowcolor{methodrowone}
Procedural navigation & Procedure dynamics and action-conditioned simulation & Mechanical thrombectomy navigation and procedure-state simulation \cite{robertshaw2025world}. \\
\rowcolor{methodrowtwo}
 & Embodied probe-motion planning & Echocardiography probe motion planning and ultrasound procedure guidance \cite{yue2025echoworld,jiang_cardiac_2024}. \\
\midrule
\rowcolor{methodrowone}
Multimodal clinical data & Cross-modal semantic alignment & Joint EHR--X-ray prediction and biomedical visual conversation \cite{elsharief_medmod_nodate,li_llava-med_2023}. \\
\rowcolor{methodrowtwo}
 & Composite medical state spaces & Unified multimodal medical understanding, tumour-state simulation and perception--dynamics--planning systems \cite{jiang2025hulu,team_lingshu_2025,yang2025medical,ding_clarity_2026,wang_brain-wm_2026}. \\
\midrule
\rowcolor{methodrowone}
Omics and cellular systems & Structure-constrained cellular state representation & Transcriptome representation learning and cellular state modelling \cite{litman_genejepa_nodate}. \\
\rowcolor{methodrowtwo}
 & Perturbation-conditioned cellular and organoid dynamics & Virtual-cell, transcriptome-response and virtual-organoid simulation \cite{chuai_towards_nodate,zhang2026lingshu,bai2026artificial}. \\
\botrule
\end{tabular}
\end{table*}

The first capability is patient-state construction from observations. Clinical observations are partial measurements rather than the state itself. In a world-model formulation, a patient state is not merely a feature vector for a downstream classifier. It should encode variables that are sufficient, or at least useful, for predicting future transitions. In medicine, this includes anatomical structure, disease burden, physiology, temporal context, patient identity, uncertainty and relevant treatment history. The state may be explicit, as in mechanistic simulators and some digital twins, or latent, as in neural sequence models and multimodal foundation models \cite{miotto_deep_2016,moor_foundation_2023,caro_brainlm_2023,yue2025chexworld}.

The second capability is clinical dynamics modelling. Dynamics modelling asks how states change over time and, when specified, under candidate interventions or control signals. In clinical settings, dynamics can be expressed through event sequences, imaging trajectories, physiological curves, tumour growth, molecular perturbation responses or surgical scene evolution. These dynamics can be learned from data, specified through mechanisms or built as a hybrid of both \cite{kraljevic_foresightgenerative_2024,yang2025medical,wang_brain-wm_2026,chuai_towards_nodate}. Importantly, dynamics models should represent uncertainty and compounding errors, because clinical rollouts are rarely deterministic and long-horizon predictions can become unreliable.

The third capability is intervention decision support. A world model becomes clinically meaningful when simulated rollouts can be compared to guide intervention choice, monitoring decisions or procedural control. This may involve estimating counterfactual outcomes, ranking treatment options, optimizing sequential policies or controlling embodied systems such as robotic instruments and imaging probes. In all cases, the model must connect candidate interventions or procedural actions to plausible future states and then compare those futures against outcomes or clinical objectives \cite{rubin2005causal,komorowski_artificial_2018,xu2026meddreamer,zhang2026world}.

This three-part definition also gives the topology of the roadmap. Existing work does not progress along a single line from representation to simulation to decision-making. Instead, it occupies three interacting capability streams. Some studies mainly build transition-ready patient states, some simulate future trajectories and some use predicted futures or learned policies to support intervention decisions. More mature medical world models arise where these streams are composed. Composition may be loose, as in serial pipelines that pass simulated futures to a decision module. It may be tighter, as in shared state-space systems that use one representation for state construction, dynamics and action-conditioned prediction. It may also approach a closed loop in which new observations update the state, candidate actions are rolled out and a planner compares future trajectories before the next intervention \cite{yang2025medical,ding_clarity_2026,wang_brain-wm_2026,he2025surgworld}.

With these axes in place, the roadmap can be read as a map of where each study sits in the simulator loop rather than as a simple chronology. Figure~\ref{fig:roadmap-taxonomy} visualizes the temporal development of the three capability streams and the emergence of composite systems. Table~\ref{tab:modality-architecture-applications} expands the same logic across modalities, method families and representative applications. Together, they organize the Review around five questions: what makes a patient state useful for world modelling (Section~\ref{sec:patient-state}), what kinds of clinical futures can be simulated (Section~\ref{sec:clinical-futures}), how simulated futures can support intervention planning (Section~\ref{sec:intervention-planning}), how partial capabilities can be composed into perception--dynamics--planning systems (Section~\ref{sec:closed-loop}) and what challenges must be solved before plausible rollouts can become clinically useful simulators (Section~\ref{sec:evaluation}).

\section{What makes a patient state useful for world modelling?}
\label{sec:patient-state}

\begin{figure}[!t]
\centering
\includegraphics[width=\textwidth]{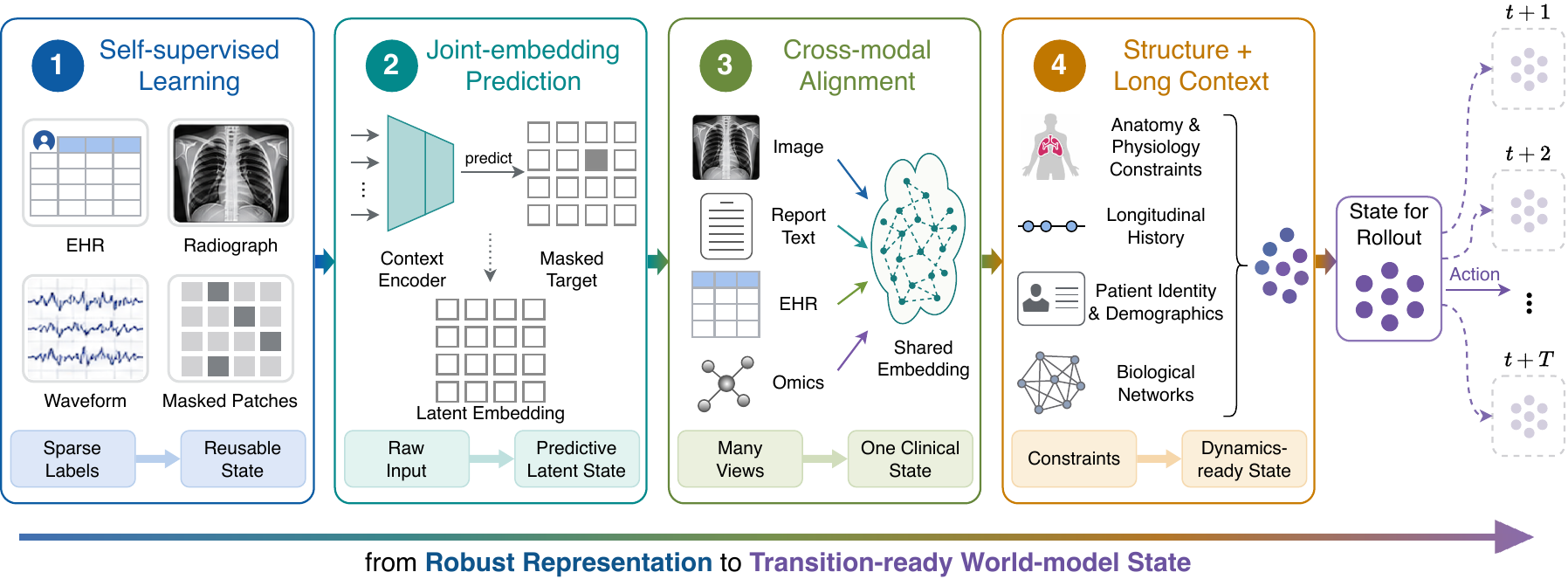}
\caption{Transition-ready patient states. Patient-state construction can be viewed as a progression from robust self-supervised representations, to predictive latent states, cross-modal clinical state spaces and structurally constrained states that are suitable for rollout in a medical world model.}
\label{fig:patient-state-progression}
\end{figure}

Medical world models begin with state construction. Raw clinical observations are heterogeneous, incomplete and noisy. A patient may be represented by images, laboratory values, diagnoses, medications, waveforms, reports, genomic measurements and physician notes. The first challenge is therefore to transform these observations into a latent state that can support future simulation. Unlike conventional representation learning, the relevant question is not only whether the representation improves a static downstream task, but whether it is transition-ready. This section follows four increasingly demanding requirements: the state should be learnable from weakly labelled clinical data, predictive in latent space, aligned across modalities and structured enough to support later dynamics (Fig.~\ref{fig:patient-state-progression}).

\subsection{Self-supervised state representation learning}
\label{subsec:self-supervised-state}

Self-supervised learning provides the first route to patient-state construction because clinical labels are expensive, sparse and often task-specific. These methods create supervision from the data themselves, for example by denoising corrupted inputs, predicting masked events or signals, matching neighbouring views or recovering held-out features. Early EHR work made this shift visible: Deep Patient used stacked denoising autoencoders to learn general-purpose patient representations from large-scale EHRs, showing that patient histories could be organized in an unsupervised latent space \cite{miotto_deep_2016}. More recent work extends the same principle to other observations. BrainLM learns masked-prediction representations of brain activity recordings, EEG2Rep constructs EEG representations from informative masked inputs, and CheXWorld explores world-model-style radiograph pretraining for transferable chest X-ray features \cite{caro_brainlm_2023,mohammadi_foumani_eeg2rep_2024,yue2025chexworld}. In this first step, the state is valuable because it is reusable and robust under missingness, noise and limited labels, but it may still be optimized mainly as a static embedding.

\subsection{Latent state learning by joint-embedding prediction}
\label{subsec:jepa-state}

Joint-embedding predictive architectures make the next step more explicit: they ask what level of the patient state should be predicted. Instead of reconstructing raw pixels or signals, a context encoder predicts the embedding of masked, future or semantically related content, and the prediction is compared with a target embedding in representation space. This matches the broader argument that useful world models should learn abstract variables that support prediction and control rather than low-level reconstruction alone \cite{lecun_path_nodate,assran2023self}. In medical settings, Brain-JEPA applies this logic to fMRI dynamics, EchoJEPA to echocardiography, Weimann et al. explore JEPA pretraining for ECG classification, EEG-VJEPA adapts V-JEPA to EEG signals, JETS addresses behavioural time series and JEPA-style surgical video models target procedure states \cite{dong_brain-jepa_2024,munim2026echojepa,weimann_self-supervised_2025,hojjati_video_2026,xie_jets_nodate,shah_vision_nodate}. At the molecular scale, GeneJepa applies the same principle to single-cell transcriptomes by predicting latent representations of masked gene sets from visible gene-expression context, yielding cellular states useful for perturbation and drug-response readouts \cite{litman_genejepa_nodate}. The point is not that these systems already simulate full clinical futures, but that their training objective makes the learned state more predictive, compact and compatible with later dynamics modelling.

\subsection{Cross-modal semantic alignment and general medical state spaces}
\label{subsec:cross-modal-state}

The third requirement is semantic alignment across clinical views. A patient state is rarely defined by one sensor. Images, reports, structured EHRs and molecular measurements are different projections of the same evolving patient, so a useful state space should preserve correspondences among them. CT2Rep and RadZero3D illustrate how 3D imaging states can be connected with reports or language-guided interpretation, whereas MedMod, LLaVA-Med, Hulu-Med and Lingshu show how EHRs, radiographs, volumes, videos, text and instruction-following can be organized into shared medical representation spaces \cite{hamamci_ct2rep_2024,park_radzero3d_nodate,elsharief_medmod_nodate,li_llava-med_2023,jiang2025hulu,team_lingshu_2025}. These systems are not necessarily world models, but they provide the cross-modal state spaces needed by composite systems that connect image states, clinical context and treatment decisions, such as CLARITY and Brain-WM \cite{ding_clarity_2026,wang_brain-wm_2026}.

\subsection{Structure-constrained and long-context state modelling}
\label{subsec:structured-state}

The final requirement is structural validity. In imaging, a useful state should respect anatomical organization, lesion geometry and patient-specific variation. X-WIN and Xray2Xray, for example, use predictive sensing or cross-view modelling to make 2D radiograph states more aware of volumetric context \cite{yang2025x,yang_xray2xray_2025}. In ultrasound, structure-aware pretraining constrains representations across views and probe positions, making the state more useful for later guidance and planning \cite{jiang2024structure}. In physiological time series, a state should respect irregular sampling, temporal continuity and individual baselines. In molecular and cellular systems, it should respect perturbation structure and biological constraints. Structure-constrained representations, time-aware sequence encoders and anatomically informed models therefore help ensure that a latent state is not only compact but also meaningful for later dynamics.

Longitudinal foundation models make this requirement explicit. Time- and person-sensitive retinal modelling, for example, treats disease prediction and risk stratification as functions of both patient identity and temporal change rather than isolated fundus photographs \cite{wang_time_2026}. Such models are not complete world models, but they move representation learning closer to patient-state dynamics.

The implication for medical world models is straightforward: static embeddings are not enough. A transition-ready patient state should preserve information needed to advance the patient through time and to evaluate how interventions or procedural actions may alter that trajectory. This makes state representation an active part of world modelling rather than a detachable pretraining stage.

\FloatBarrier

\section{What kinds of clinical futures can be simulated?}
\label{sec:clinical-futures}

\begin{figure}[!t]
\centering
\includegraphics[width=\textwidth]{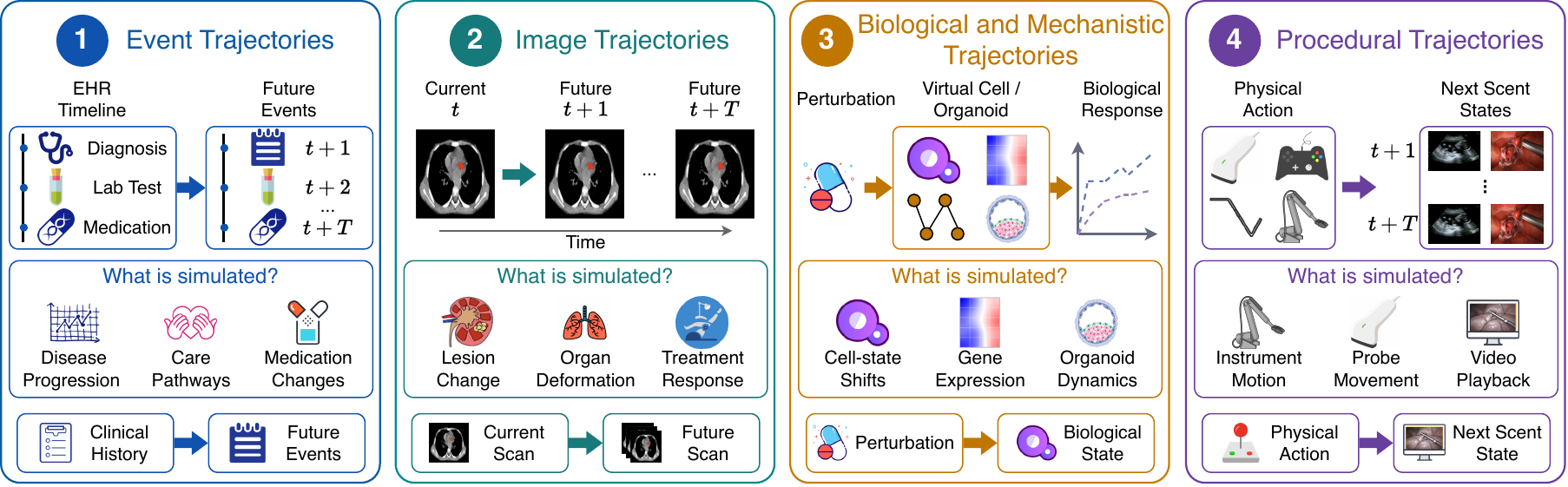}
\caption{Clinical future spaces for medical world models. Dynamics models can simulate different kinds of futures, from symbolic EHR event trajectories and visible imaging trajectories to biological and mechanistic trajectories and action-conditioned procedural trajectories. Each future space requires a different transition mechanism and exposes different evaluation risks.}
\label{fig:future-spaces}
\end{figure}

Dynamics modelling is the component that most clearly separates medical world models from static medical AI. A classifier maps observations to labels. A world model maps states to possible futures. In medicine, the simulated future is not a single object. It may be a recorded care timeline, a visible anatomical change, a biological or mechanistic state change or an action-conditioned procedural scene (Fig.~\ref{fig:future-spaces}). This section therefore organizes clinical futures by the object being advanced. The sequence moves from clinical traces that are readily observed, to visible disease states, to underlying biological mechanisms, and finally to futures driven by physical control inputs. Each future space has a different strength and a different failure mode. Event trajectories are available at scale but entangle disease with care processes. Image trajectories are inspectable but may be visually plausible without being clinically valid. Biological and mechanistic trajectories move closer to mechanism but are harder to observe and validate. Procedural trajectories make actions explicit but must satisfy strict anatomical, physical and safety constraints.

\subsection{Event trajectories}

In EHR settings, the future is naturally recorded as a sequence of diagnoses, medications, procedures, laboratory tests and visits. This makes event trajectories the most accessible longitudinal substrate for medical world models. They are discrete, irregularly sampled and available at scale, and they contain traces of disease progression, treatment response and clinical deterioration. The modelling step is to turn an observed history into a distribution over plausible next events or longer patient timelines. Early models such as Disease Trajectory Maps and Doctor AI treated patient histories as temporal objects rather than static feature tables, enabling disease progression and next-visit events to be modelled explicitly \cite{schulam_disease_2016,choi_doctor_2016}. More recent generative approaches extend this logic from next-event prediction to patient timeline generation. Foresight uses a generative pretrained transformer to model large-scale EHR timelines and generate future clinical events over longer horizons, while Curiosity models and longitudinal EHR world-model paradigms emphasize that the patient should be represented as an evolving state rather than a moving document \cite{kraljevic_foresightgenerative_2024,waxler_generative_2025,adam2026patient}. EHRWorld further couples patient-centric state consistency with long-horizon dynamics \cite{mu_ehrworld_2026}.

The strength of event trajectories is also their weakness. They make clinical history simulatable, but the recorded future is not purely biological. It mixes disease progression with care pathways, coding practice, access to care, monitoring intensity and clinician behaviour. A medical world model built on EHR events must therefore distinguish, as far as possible, between patient-state transitions and observation-process artefacts.

\subsection{Image trajectories}

Image trajectories move from recorded care traces to visible disease states. In longitudinal imaging, futures appear as changes in MRI, CT, radiographs or ultrasound. Models may simulate lesion growth, organ deformation, tumour response or post-treatment anatomy. Their advantage is interpretability: a future scan lets clinicians inspect what the model believes will change and where that change will occur. The central modelling question is which transition mechanism moves the current image state forward. GANs, diffusion models, latent diffusion and flow matching have been used to model Alzheimer's disease progression, future brain MRI, treatment-aware glioma evolution, post-treatment glioblastoma response and patient-specific imaging trajectories \cite{zhao_prediction_2021,puglisi_brain_2025,liu_treatment-aware_2025,chen_learning_2026,leclercq_pre_2026}. Seq2Morph, MRI contrast-enhancement kinetics models and DeepPrognosis show complementary routes in which temporal deformation, contrast dynamics and dynamic CT evidence represent disease evolution and treatment-relevant future states \cite{lee_seq2morph_2023,kong2026mri,yao_deepprognosis_2021}. ImageFlowNet and related approaches emphasize irregular time intervals and multiscale lesion changes \cite{liu_imageflownet_2025}. CRONOS pushes this problem into continuous-time 3D sequence-to-image forecasting by learning a spatiotemporal velocity field that transports multiple context volumes toward a target volume at an arbitrary timestamp \cite{disch2025cronos,disch2025temporal}.

The difficulty is that visual plausibility is an incomplete medical criterion. A simulated future scan can support explanation and treatment-response reasoning only if it is temporally plausible, clinically calibrated and sensitive to interventions. Otherwise, image generation risks becoming visual extrapolation rather than disease simulation.

\subsection{Biological and mechanistic trajectories}

A third future space lies below the level of clinical records and visible anatomy. Here, the simulated object is a biological or mechanistic state. At cellular and molecular scales, the future may be a perturbation-induced shift in gene expression, cell state or drug response. At organoid, tissue and organ scales, the future may be a mechanism-constrained trajectory of cellular interaction, transport, glucose regulation, haemodynamics, tumour burden or organ function. Recent cellular world models make this direction explicit. Lingshu-Cell uses a masked discrete diffusion model to learn transcriptomic state distributions and conditionally simulate whole-transcriptome responses under new combinations of cell identity, donor identity and perturbation \cite{zhang2026lingshu}. AlphaCell represents a complementary virtual-cell route by evolving latent cell states in a virtual cell space to model perturbation-induced cellular dynamics \cite{chuai_towards_nodate}. At organoid and tissue scales, artificial intelligence virtual organoids frame organoid-scale digital twins as executable biological systems that combine virtual cells, multimodal longitudinal readouts, virtual instruments and hybrid mechanistic modules for assay, perturbation and drug-response simulation \cite{bai2026artificial}. Mechanistic simulators and digital twins provide a more explicitly specified route at organ and system scales. The UVA/PADOVA type 1 diabetes simulator, in silico clinical trial frameworks and the Virtual Physiological Human encode biological mechanisms and treatment responses explicitly \cite{man_uvapadova_2014,pappalardo_silico_2019,viceconti_virtual_2016}.

The strength of this future space is mechanistic relevance. It can ask how a perturbation changes a biological system rather than only how a record or image changes. Its difficulty is evidence. Longitudinal biological readouts are sparse, expensive and heterogeneous, and validation often requires experimental or physiological constraints. Medical rollouts in this space must therefore satisfy biological and mechanistic plausibility, not only observational continuity.

\subsection{Procedural trajectories}

Procedural trajectories make the action variable most concrete. In embodied medical AI, the future is a procedure state generated under a physical control input: probe motion, instrument movement, catheter direction or manipulation strategy. This makes the world-model formulation especially natural, because the model can connect perception, action and next-state prediction within the same procedural scene. Surgical video world models, robotic suturing models, surgical grasping models, ultrasound probe guidance systems, autonomous robotic ultrasound and endovascular navigation models aim to predict how a scene or anatomy will change under such actions \cite{koju_surgical_2026,turkcan2025towards,lin_world_2024,yue2025echoworld,jiang_cardiac_2024,bi_autonomous_2024,robertshaw2025world}. He et al. extend this logic towards learning surgical robot policies from videos through world modelling \cite{he2025surgworld}.

Compared with disease progression, the local transition in procedural settings can be easier to define because actions are observable and physically grounded. This does not make the clinical problem simple. Simulated actions must respect anatomy, safety margins, operator skill, device physics and workflow realities, and short-horizon scene prediction still needs to be connected to meaningful clinical outcomes.

Across these future spaces, the central question is not only whether a model predicts accurately. It is what kind of future the model can simulate, what makes that future clinically useful, whether the future is conditioned on interventions or procedural actions, how uncertainty is represented and how errors compound during rollout.

\FloatBarrier

\section{How can simulated futures support intervention planning?}
\label{sec:intervention-planning}

\begin{figure}[!t]
\centering
\includegraphics[width=\textwidth]{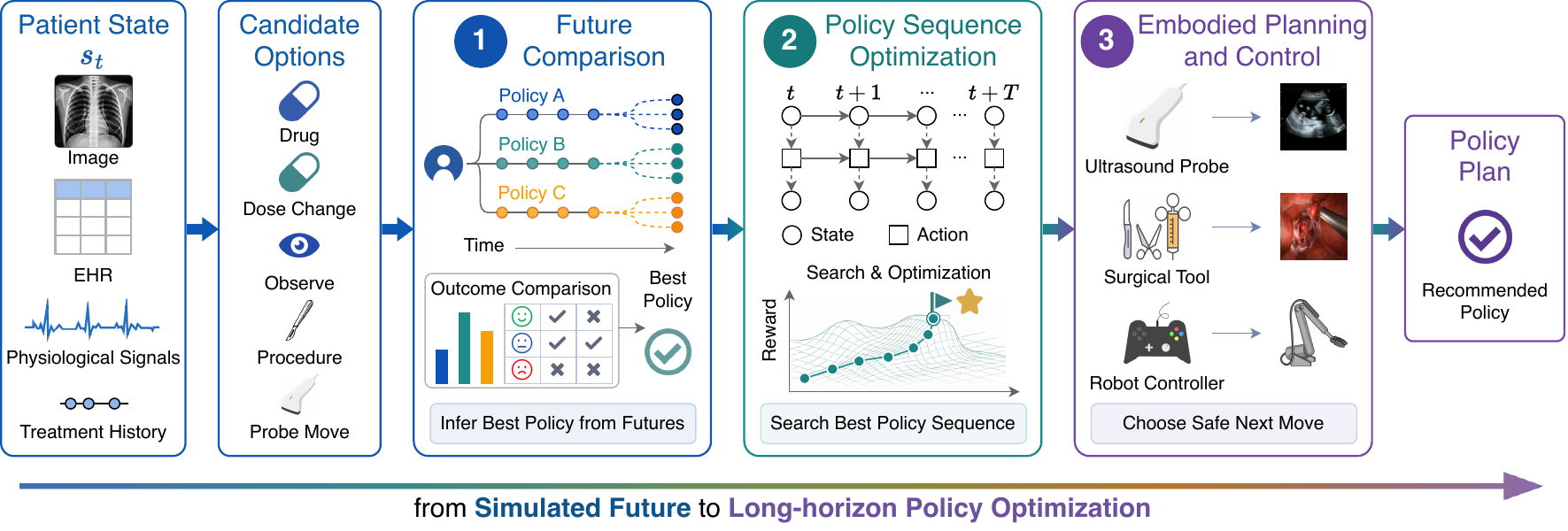}
\caption{Simulated futures for intervention planning. Medical world models support three decision interfaces for intervention: simulated outcome evidence, which combines risk scoring with counterfactual outcome comparison; policy optimization, which actively searches sequential treatment strategies; and embodied planning and control, which selects safe physical actions in procedural environments.}
\label{fig:intervention-planning}
\end{figure}

Prediction becomes clinically meaningful when it informs intervention. A medical world model should therefore do more than extrapolate disease trajectories. It should help compare possible futures under alternative interventions or procedural actions. This shifts the problem from forecasting to planning across three increasingly active decision interfaces: simulated outcome evidence, sequential policy optimization and embodied planning and control (Fig.~\ref{fig:intervention-planning}).

\subsection{Simulated outcome evidence}

Individualized risk and treatment-effect assessment provide the most direct form of simulated outcome evidence. Models such as DeepSurv and DeepPrognosis show how patient features, imaging evidence and treatment context can be converted into decision-relevant risk estimates or survival predictions \cite{katzman_deepsurv_2018,yao_deepprognosis_2021}. These systems can assist decisions, but a risk score alone does not fully answer the intervention question. Counterfactual outcome methods go further by estimating what would happen if a different treatment trajectory were chosen. Adversarially balanced representations, SyncTwin and G-Net estimate potential outcomes under alternative treatment paths, making the comparison among interventions more explicit \cite{bica_estimating_2020,qian_synctwin_nodate,li_g-net_nodate}. The relevant question is often not whether a patient is high risk, but which intervention is expected to improve the future trajectory.

Medical world models extend this idea by inserting generative simulation into the decision chain. MeWM uses candidate treatment generation, post-intervention tumour-state simulation and outcome evaluation to compare strategies \cite{yang2025medical}. CLARITY models context-aware disease trajectories in latent space and converts treatment-conditioned simulations into decision evidence \cite{ding_clarity_2026}. In both cases, the simulated future becomes an intermediate object for reasoning rather than a final prediction.

\subsection{Policy optimization}

Sequential policy optimization broadens intervention planning from comparing candidate interventions to learning a strategy over time. Reinforcement learning formulates care as a sequence of states, actions and rewards. The Artificial Intelligence Clinician used ICU data to learn treatment strategies for sepsis, comparing learned policies with clinician decisions \cite{komorowski_artificial_2018}. Continuous state-space approaches for sepsis and newer offline reinforcement learning methods for acute kidney injury further illustrate how clinical policy learning can be grounded in patient dynamics \cite{raghu_continuous_nodate,zhang2026world}. MedDreamer embodies the idea of latent imagination by learning EHR dynamics and evaluating candidate strategies through imagined futures \cite{xu2026meddreamer}.

Policy optimization highlights the promise and risk of medical world models. A learned simulator can improve sample efficiency and reduce the need for unsafe exploration, but it can also amplify bias if the simulator is wrong. Offline clinical data are confounded by physician behaviour, institutional practice and treatment-selection bias. Any policy derived from a learned world model must therefore be constrained by uncertainty, clinical plausibility and prospective validation.

\subsection{Embodied planning and control}

Embodied planning extends intervention planning to procedural environments where actions are physical and immediate. In ultrasound, surgery and endovascular navigation, systems such as autonomous robotic ultrasound, EchoWorld, Cardiac Copilot, surgical grasping models, He et al.'s surgical policy-learning framework and robotic navigation world models learn how probe movements, instrument actions or catheter trajectories change future observations \cite{bi_autonomous_2024,yue2025echoworld,jiang_cardiac_2024,lin_world_2024,he2025surgworld,robertshaw2025world}. Here, planning is not an abstract recommendation but a control problem. The action space, reward structure and safety constraints are defined by anatomy, device mechanics and procedural goals.

Despite their differences, treatment recommendation and embodied control share the same world-model logic: represent the current state, simulate intervention- or action-conditioned futures and choose interventions or procedural actions by comparing those futures. This shared logic is what allows medical world models to connect clinical prediction, counterfactual inference, reinforcement learning and robotics under one framework.

\FloatBarrier

\section{Towards closed-loop perception-dynamics-planning systems}
\label{sec:closed-loop}

\begin{figure*}[t]
\centering
\includegraphics[width=\textwidth]{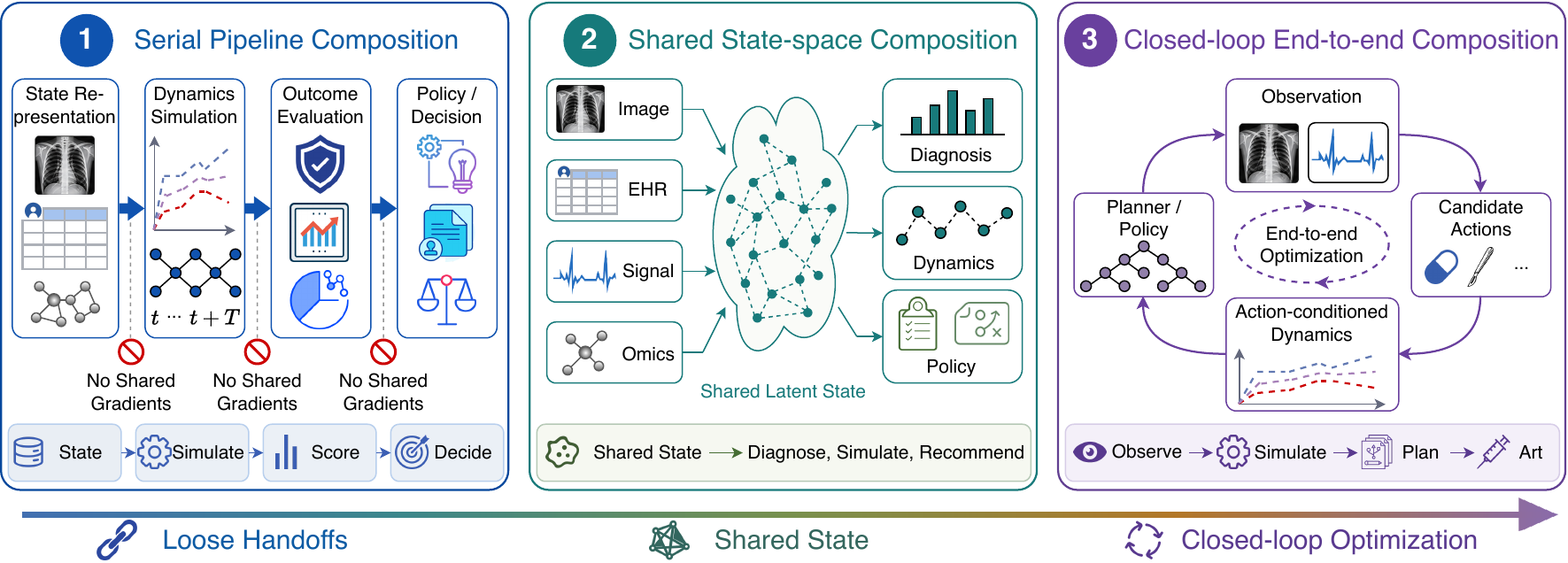}
\caption{Compositional patterns for medical world models. Serial pipeline composition connects independently optimized modules through explicit handoffs, making implementation straightforward but allowing errors to accumulate downstream. Shared state-space composition makes state representation, dynamics simulation and decision modules operate on a common representation, improving interface alignment. Closed-loop end-to-end composition optimizes the perception--dynamics--planning loop jointly, which can reduce interface error but is hardest to train and is currently most mature in procedural settings.}
\label{fig:composition-patterns}
\end{figure*}

Most current systems solve only part of the medical world-model loop. Some learn states without modelling dynamics. Some model future trajectories without supporting decisions. Some recommend interventions without an explicit simulator. The key question for composite medical world models is therefore architectural: how are disease-state understanding, disease dynamics modelling and intervention decision support wired into one system? Existing work can be organized into three compositional patterns: serial pipeline composition, shared state-space composition and closed-loop simulator composition (Fig.~\ref{fig:composition-patterns}).

\subsection{Serial pipeline composition}
\label{subsec:serial-pipeline-composition}

Serial pipeline composition links capabilities through explicit handoffs between modules. The state interface may be an image, an EHR history, a tumour mask or a learned embedding. What matters is that a simulator is explicitly connected to an action or decision module. MeWM is the clearest example. It composes a vision-language policy model that proposes candidate treatments, a tumour generative model that simulates post-intervention tumour states and an inverse-dynamics survival module that evaluates those simulated outcomes to select a treatment plan \cite{yang2025medical}. CLARITY follows a related chain in latent space: patient-specific clinical context and time intervals initialize a treatment-conditioned disease trajectory, and a prediction-to-decision module translates the latent rollout into actionable treatment evidence \cite{ding_clarity_2026}. In EHR-based decision settings, MedDreamer uses a latent world model to simulate irregular patient trajectories and then trains a policy on a mixture of real and imagined experience for treatment recommendation \cite{xu2026meddreamer}. WME-ORL composes three functions in a more explicitly policy-centred chain. An ensemble FNO--Transformer world model predicts patient-state dynamics and AKI-stage progression. The resulting predictions and uncertainty estimates modify stage-aware implicit Q-learning for sequential intervention optimization, while clinical-rule penalties constrain unsafe or distribution-shifted actions \cite{zhang2026world}. It therefore combines patient-state and disease-stage modelling, dynamics-based trajectory evaluation and intervention optimization for decisions such as fluid balance, diuretics and renal replacement therapy. Serial composition is attractive because the functions are inspectable: one can ask what state was used, what future was generated and how the future was scored. Its limitation is that the modules may still be optimized separately, so errors in state estimation or rollout can be handed downstream to the decision module.

\subsection{Shared state-space composition}
\label{subsec:shared-state-composition}

Shared state-space composition tightens the interface by making simulation and decision components operate on the same patient, cellular or procedural state space. The shared object is not always a single latent vector. It may be a tumour-progression representation, a virtual-cell manifold or a surgical feature state. What matters is that the same state space supports three interfaces: state representation, dynamics simulation and prediction for intervention decisions or procedural actions. Brain-WM is the clearest multitask example. It encodes treatment history, demographics and multimodal MRI into a shared spatiotemporal tumour state. A Y-shaped Mixture-of-Transformers then uses this state for both autoregressive next-step treatment prediction and flow-based future MRI generation, while multi-timepoint mask alignment anchors intermediate representations to anatomically grounded tumour structures \cite{wang_brain-wm_2026}. AlphaCell offers a particularly explicit state-space design. A genome-wise encoder constructs a continuous Virtual Cell Space, a decoder maps latent cell states back to genome-wide expression profiles and a perturbation-conditioned flow model evolves those states within the same manifold \cite{chuai_towards_nodate}. Shah et al.'s SurgWorld is a weaker but informative boundary case. It uses a frozen cataract video encoder as the surgical state space, a latent action tokenizer to infer action primitives from frame-to-frame motion and an action-conditioned predictor to model next-state features for step recognition and anticipation \cite{shah2026learning}. These systems move beyond a pipeline in which state representation, dynamics simulation and decision modules are arranged as separate steps. Instead, the functions read from, update or branch from a shared state space. The remaining limitation lies at the decision interface. Treatments, perturbations or surgical actions are usually modelled as observed variables, inferred tokens or supervised targets, not as candidate decisions to be rolled out, compared and optimized through counterfactual planning.

\subsection{Closed-loop simulator composition}
\label{subsec:closed-loop-simulator-composition}

The strongest form of composition treats actions not only as conditions for prediction, but as candidate choices inside a feedback loop. New observations update the patient or procedure state. Candidate actions query an action-conditioned dynamics model, and simulated futures are scored and compared by a planner. The selected or observed intervention changes the real patient or procedure, and the next observations reinitialize the loop. This closed-loop form remains closer to a target architecture than to a mature clinical reality. Current examples come mainly from embodied and procedural settings because their loops are physically local and densely observed. EchoWorld and Cardiac Copilot compose anatomical state representation, probe-motion dynamics and real-time guidance for echocardiography \cite{yue2025echoworld,jiang_cardiac_2024}. Robertshaw et al.'s mechanical-thrombectomy navigation framework uses TD-MPC2 to connect patient-specific vascular environments with long-horizon catheter control \cite{robertshaw2025world}. He et al.'s SurgWorld composes a generative surgical world model, inverse dynamics and synthetic video-action data to improve downstream surgical robot policy learning \cite{he2025surgworld}. These systems approximate closed-loop composition because the state is often a visible scene or anatomical geometry, the action is a probe, catheter or instrument displacement, and the transition can be evaluated as a change in view, position or procedure state over a short horizon. This is more tractable than longitudinal disease simulation, where actions are higher-level clinical interventions, effects may be delayed, outcomes are partly unobserved and trajectories are confounded by care practice. The broader challenge is to transfer this closed-loop form from local motion control to patient-level disease simulators, where simulated futures must support treatment planning under uncertainty rather than only procedural navigation.

This compositional taxonomy clarifies the current direction of the field. Medical world models are moving from isolated components to systems that coordinate multiple capabilities: serial pipelines connect simulation to decision evidence, shared state-space systems align state representation, simulation and prediction for intervention decisions around one representation and procedural systems begin to connect prediction with action. The remaining gap is closed-loop reliability for disease-level simulation. A clinically useful composite model must make its interfaces explicit: what state is being simulated, what intervention or action is being conditioned on, what future is generated, how that future is evaluated and how new observations update the next decision. This gap between present composite systems and patient-level clinical simulators motivates the evaluation and translation challenges discussed below.

\section{Open challenges: from plausible rollouts to clinically useful simulators}
\label{sec:evaluation}

Medical world models are beginning to produce plausible rollouts, but plausibility is not the same as clinical usefulness. A future scan may look realistic, an EHR timeline may follow common care patterns and an intervention-conditioned trajectory may match observed associations, yet none of these outputs alone proves that the model has learned a reliable simulator of patient-state evolution. The central question for the next stage of the field is therefore not only how to generate futures, but what evidence would make those futures useful for clinical reasoning, intervention comparison and workflow-integrated decision support \cite{qazi_beyond_2025}.

\subsection{Algorithmic challenges across the world-model loop}

State adequacy is the first pressure point. A model cannot predict clinically meaningful futures if the input state omits variables that determine future transitions. Disease stage, anatomical burden, physiology, treatment history, prior response, comorbidities, timing, measurement practice and institutional context may all change the expected rollout. If key determinants are absent, apparently plausible trajectories may be driven by hidden confounding or unobserved clinical context. If irrelevant details dominate the state, the model may overfit local patterns without improving transition validity. Evaluation should therefore ask whether the learned state is sufficient for intervention- or action-conditioned prediction: whether adding omitted clinical context changes the rollout, whether uncertainty increases when information is missing and whether ablations of treatment history, timing or disease stage alter the simulated future in clinically sensible ways \cite{wang_brain-wm_2026}.

Dynamics modelling brings a different difficulty. In medicine, the future is rarely a single path. Disease progression, treatment response, complications, clinician behaviour, monitoring frequency and patient adherence can all create distinct but plausible trajectories from the same apparent state. A useful simulator should not collapse these alternatives into one average future. It should distinguish uncertainty that can be reduced by better state measurement from uncertainty that reflects biological heterogeneity, stochastic events or genuinely underdetermined clinical context. This requires evaluating whether models generate diverse but clinically coherent futures, whether uncertainty grows over longer rollouts, whether rare but consequential outcomes remain visible and whether different candidate interventions produce distinguishable future distributions rather than cosmetically different samples \cite{kraljevic_foresightgenerative_2024,shmatko_natural_2025}.

Intervention planning adds a causal problem. Conditioning on an observed treatment is not the same as simulating an intervention. Clinical interventions are confounded by severity, clinician judgement, access to care, institutional protocols and patient preference. A model trained naively on observational trajectories may learn how clinicians behaved for particular patients rather than how those patients would have evolved under alternative interventions. Medical world models that support intervention planning will therefore need target-trial emulation, causal representation learning, counterfactual validation, sensitivity analyses for treatment-selection bias and prospective or quasi-prospective testing before their intervention-conditioned rollouts can be treated as decision evidence \cite{rubin2005causal,hernan2010causal,bica_estimating_2020}.

Closed-loop composition turns these issues into a system-level problem. Many current systems combine a state encoder, a trajectory generator and a decision head, but do not yet show that the full perception--dynamics--planning loop remains reliable when embedded in clinical workflows. A clinically useful simulator should make its state assumptions explicit, communicate uncertainty, expose alternative futures for inspection and support human oversight rather than opaque automation. The next stage is not merely to build larger medical generative models, but to build clinically testable simulators whose states are sufficient, transitions are intervention-sensitive, futures are uncertain but calibrated and recommendations remain inspectable by clinicians \cite{laubenbacher_digital_twins_2024,sadee_medical_digital_2025,luo_clinical_environment_2026}.

\subsection{Data, FAIR infrastructure and evaluation}

The remaining challenges are infrastructural. The limiting resource for medical world models is not simply the number of patients, but the availability of records that connect a current state, an intervention or action, a time interval and a subsequent state. In abstract form, useful evidence should approximate $(s_t,a_t,\Delta t)\rightarrow s_{t+1}$, together with downstream outcomes and measurement context. Many clinical datasets instead contain isolated snapshots, sparse follow-up measurements, outcome labels without intermediate disease states or longitudinal images without sufficiently detailed intervention timing, dose and treatment context. Models trained on such data may learn associations between observations and outcomes, but lack the temporal evidence needed to learn how patient states move across disease stages and under clinical interventions \cite{kraljevic_foresightgenerative_2024,shmatko_natural_2025,mu_ehrworld_2026}.

These data requirements also raise FAIR (findable, accessible, interoperable and reusable) and governance challenges. Clinical simulators need longitudinal data that are findable across institutions, accessible under appropriate governance, interoperable across modalities and reusable for external validation without exposing patients to avoidable privacy risks. For medical world models, FAIR infrastructure is not only a data-management ideal. It determines whether state definitions, action labels, timing variables, outcome measures and missingness mechanisms can be compared across sites. Future studies will therefore need privacy-preserving data pipelines, transparent cohort and action definitions, institution-level generalization tests and accountability mechanisms for cases in which simulated futures influence care.

Evaluation should then be trajectory-level rather than snapshot-level. Different future spaces require different standards. EHR rollouts should be judged by event distributions, temporal ordering, missingness patterns and care-process artefacts. Image rollouts require anatomical consistency, lesion continuity, treatment sensitivity and calibration against follow-up scans. Biological and mechanistic simulations require agreement with continuous measurements, perturbation responses, mechanistic constraints and individual baselines. Procedural rollouts require safety, controllability, device realism and robustness under distribution shift. Across these settings, evaluation should move beyond whether a generated future is plausible at one time point and ask whether the full trajectory is coherent, clinically meaningful and useful for comparing actions at the horizon that matters \cite{luo_clinical_environment_2026,chen_surgeons_world_2025}.

\section{Conclusion}
\label{sec:conclusion}

Medical world models offer a way to reframe the next stage of medical AI. The aim is not only to improve diagnosis or prediction, but to build systems that can represent medical states, model clinical dynamics and guide intervention policies through simulated futures. This Review has developed this framing as a roadmap rather than as a single architecture. It defines medical world models as clinical simulators, translates the general world-model formulation into three medical capabilities and organizes fragmented work across EHR modelling, imaging, biological simulation, treatment-effect estimation, reinforcement learning, procedural control and digital twins.

This roadmap makes three contributions to the field. It clarifies that state representation is valuable for world modelling only when it preserves information needed for future transitions. It distinguishes several forms of clinical futures, including event sequences, image trajectories, biological and mechanistic trajectories and procedural scene evolution. It also separates intervention support into outcome prediction, counterfactual reasoning, policy optimization and embodied control. These distinctions make it possible to compare methods that are often discussed in separate literatures, and to identify where current systems remain partial components rather than complete perception--dynamics--planning loops.

The same framing also defines the work still needed. Clinically useful simulators will require longitudinal multimodal benchmarks, stronger causal and mechanistic grounding, intervention-aware training objectives, trajectory-level evaluation, calibrated uncertainty and governance mechanisms for high-stakes use. Their value should not be measured by whether generated futures look plausible in isolation, but by whether those futures are coherent, action-sensitive, inspectable and useful for comparing interventions at clinically meaningful horizons.

The medical world-model view therefore provides both a synthesis and a caution. It encourages the integration of representation learning, generative dynamics, causal inference, reinforcement learning and embodied AI. At the same time, it makes clear that larger models alone will not produce reliable clinical simulators. Progress will depend on meaningful medical states, valid dynamics, well-specified interventions, calibrated uncertainty and human oversight. Under these conditions, medical world models could shift medical AI from recognizing present disease toward reasoning about possible futures and the interventions that may change them.

\backmatter

\bibliography{bib/ref}

\FloatBarrier
\clearpage

\begin{appendices}
\renewcommand{\theHtable}{app.\arabic{table}}

\section{Key concepts in medical world models}
\label{app:key-concepts}

\begingroup
\small
\noindent\textbf{Box 2 | Key concepts in medical world models.}
\vspace{0.5em}

\begin{description}
\item[Observation] A measurement or record available to the model, such as images, EHR events, laboratory values, physiological signals, reports, molecular data or procedural video. Observations are often partial, noisy and irregularly sampled, and are used to infer patient state.
\item[Patient state] A compact representation of the current patient, inferred from observations such as images, EHRs, physiological signals, reports, molecular data or multimodal records. A useful state should preserve information needed for future transitions, not only for static prediction.
\item[Policy, intervention or action] A decision variable or concrete input that can alter future states. In medicine this includes treatment policies, doses, procedures, monitoring decisions, follow-up strategies, molecular perturbations, probe movements, robotic commands or the choice to withhold treatment.
\item[Transition or dynamics model] A model of how patient states evolve over time, commonly written as $p(s_{t+1}\mid s_t,a_t)$ when transitions are action-conditioned. Dynamics may be learned from data, specified by mechanisms or built as a hybrid.
\item[Rollout] A simulated sequence of future states generated by repeatedly applying a transition model. Rollouts can expose possible trajectories, but they also accumulate uncertainty and model error.
\item[Objective or cost] A criterion used to train predictions or judge future states, actions or trajectories. In medicine, this may reflect survival, safety, treatment response, functional outcome, complication avoidance, information gain or procedural success. In reinforcement-learning settings, it may be implemented as a reward.
\item[Counterfactual trajectory] A possible future under an alternative action or treatment path. Counterfactual trajectories are central to decision-making but require causal assumptions beyond ordinary forecasting.
\end{description}
\endgroup

\FloatBarrier

\end{appendices}

\end{document}